\documentclass[11pt]{article}
\usepackage{coling2020}
\usepackage{times}
\usepackage{url}
\usepackage{latexsym}
\usepackage{amsmath}
\usepackage{graphicx}
\usepackage{multirow}
\usepackage{subcaption}
\usepackage{makecell}
\usepackage{xcolor}
\usepackage{CJKutf8}
\usepackage{arydshln}
\usepackage{hyperref}

\DeclareMathOperator*{\argmax}{arg\,max}
\newcommand{\Lagr}{\mathcal{L}}
\newcommand{\curlU}{\mathcal{U}}

\title{Leveraging Discourse Rewards\\for Document-Level Neural Machine Translation }

\author{Inigo Jauregi Unanue$^{\textbf{1,3}}$, Nazanin Esmaili$^\textbf{1}$, Gholamreza Haffari$^\textbf{2}$, Massimo Piccardi$^\textbf{1}$\\
  $^1$University of Technology Sydney, NSW 2007, Australia\\
  $^2$Monash University, VIC 3800, Australia \\
  $^3$RoZetta Technology, NSW 2000, Australia \\
  {\tt Inigo.Jauregi@rozettatechnology.com} \\
  {\tt \{Nazanin.Esmaili,Massimo.Piccardi\}@uts.edu.au} \\
  {\tt Gholamreza.Haffari@monash.edu}}

\date{}

\begin{document}
\maketitle
\begin{abstract}
  Document-level machine translation focuses on the translation of entire documents from a source to a target language. It is widely regarded as a challenging task since the translation of the individual sentences in the document needs to retain aspects of the discourse at document level. However, document-level translation models are usually not trained to explicitly ensure discourse quality. Therefore, in this paper we propose a training approach that explicitly optimizes two established discourse metrics, \textit{lexical cohesion} (LC) and \textit{coherence} (COH), by using a reinforcement learning objective. Experiments over four different language pairs and three translation domains have shown that our training approach has been able to achieve more cohesive and coherent document translations than other competitive approaches, yet without compromising the faithfulness to the reference translation. In the case of the Zh-En language pair, our method has achieved an improvement of $2.46$ percentage points (pp) in LC and $1.17$ pp in COH over the runner-up, while at the same time improving $0.63$ pp in BLEU score and $0.47$ pp in $\mathrm{F}_{\mathrm{BERT}}$.
  
\end{abstract}

\section{Introduction}
\label{sec:introduction}

The recent advances in neural machine translation (NMT) \cite{sutskever2014,bahdanau2014,luong2015,vaswani2017attention} have provided the research community and the commercial landscape with effective translation models that can at times achieve near-human performance. However, this usually holds at phrase or sentence level. When using these models in larger units of text, such as paragraphs or documents, the quality of the translation may drop considerably in terms of discourse attributes such as lexical and stylistic consistency.

In fact, document-level translation is still a very open and challenging problem. The sentences that make up a document are not unrelated pieces of text that can be predicted independently; rather, a set of sequences linked together by complex underlying linguistics aspects, also known as the discourse \cite{maruf2019survey,jurafsky2019speech}. The discourse of a document includes several properties such as grammatical cohesion \cite{halliday2014cohesion}, lexical cohesion \cite{halliday2014cohesion}, document coherence \cite{hobbs1979coherence} and the use of discourse connectives \cite{kalajahi2012discourse}. Ensuring that the translation retain such linguistic properties is expected to significantly improve its overall readability and flow.

However, due to the limitations of current decoder technology, NMT models are still bound to translate at sentence level. In order to capture the discourse properties of the source document in the translation, researchers have attempted to incorporate more contextual information from surrounding sentences. Most document-level NMT approaches augment the model with multiple encoders, extra attention layers and memory caches to encode the surrounding sentences, and leave the model to implicitly learn the discourse attributes by simply minimizing a conventional NLL objective. The hope is that the model will spontaneously identify and retain the discourse patterns within the source document. Conversely, very little work has attempted to model the discourse attributes explicitly. Even the evaluation metrics typically used in translation such as BLEU \cite{papineni2002bleus} are not designed to assess the discourse quality of the translated documents.

For these reasons, in this paper we propose training an NMT model by directly targeting two specific discourse metrics: \textit{lexical cohesion} (LC) and \textit{coherence} (COH). LC is a measure of the frequency of semantically-similar words co-occurring in a document (or block of sentences) \cite{halliday2014cohesion}. For example, \textit{car}, \textit{vehicle}, \textit{engine} or \textit{wheels} are all semantically-related terms. There is significant empirical evidence that ensuring lexical cohesion in a text eases its understanding \cite{halliday2014cohesion}. At its turn, COH measures how well adjacent sentences in a text are linked to each other. In the following example from Hobbs \shortcite{hobbs1979coherence}:

\begin{center}	
``John took a train from Paris to Istanbul. He likes spinach.''	
\end{center}

\vspace{12pt}

\noindent the two sentences make little `sense' one after another. An incoherent text, even if grammatically and syntactically perfect, is anecdotally very difficult to understand and therefore coherence should be actively pursued. Relevant to translation, Vasconcellos \shortcite{vasconcellos1989cohesion} has found that a high percentage of the human post-editing changes over machine-generated translations involves the improvement of cohesion and coherence.

Several LC and COH metrics that well correlate with the human judgement have been proposed in the literature. However, like BLEU and most other evaluation metrics, they are discrete, non-differentiable functions of the model's parameters. Hereafter, we propose to overcome this limitation by using the well-established \textit{policy gradient} approach from reinforcement learning \cite{sutton1999,sutton2018reinforcement} which allows using any evaluation metric as a reward without having to differentiate it. By combining different types of rewards, the model can be trained to simultaneously achieve more lexically-cohesive and more coherent document translations, while at the same time retaining faithfulness to the reference translation.


\section{Related Work}
\label{sec:related}


\subsection{Document-level NMT}
\label{ssec:dl_nmt}

Many document-level NMT models have proposed taking the context into account by concatenating surrounding sentences or extra features to the current input sentence, with otherwise no modifications to the model. For example, Rios et al. \shortcite{rios2017improving} have trained an NMT model that learns to disambiguate words given the context semantic landscape by simply extracting lexical chains from the source document, and using them as additional features. Other researchers have proposed concatenating previous source and target sentences to the current source sentence, so that the decoder can observe a proper amount of context \cite{agrawal2018contextual,tiedemann2017neural,scherrer2019analysing}. Their work has shown that concatenating even just one or two previous sentences can result in a noticeable improvement. Mac{\'e} and Servan \shortcite{mace2019using} have added an embedding of the entire document to the input, and shown promising results in English-French.

Conversely, other document-level NMT approaches have proposed modifications to the standard encoder-decoder architecture to more effectively account for the context from surrounding sentences. Jean et al. \shortcite{jean2017does} have introduced a dedicated attention mechanism for the previous source sentences. Multi-encoder approaches with hierarchical attention networks have been proposed to separately encode each of the context sentences before they are merged back into a single context vector in the decoder \cite{miculicich2018document,maruf2019selective,wang2017exploiting}. These models have shown significant improvements over sentence-level NMT baselines on many different language pairs. Kuang et al. \shortcite{kuang2017modeling} and Tu et al. \shortcite{tu2018learning} have proposed using an external cache to store, respectively, a set of topical words or a set of previous hidden vectors. This information has proved to benefit the decoding step at limited additional computational cost. In turn, Maruf and Haffari \shortcite{maruf2017document} have presented a model that incorporates two memory networks, one for the source and one for the target, to capture document-level interdependencies. For the inference stage, they have proposed an iterative decoding algorithm that incrementally refines the predicted translation.

However, all the aforementioned models assume that the model can implicitly learn the occurring discourse patterns. Moreover, the training objective is the standard negative log-likelihood (NLL) loss, which simply maximizes the probability of the reference target words in the sentence. Only one work these authors are aware of \cite{xiong2019modeling} has attempted to train the model by explicitly learning discourse attributes. Inspired by recent work in text generation \cite{bosselut2018discourse}, Xiong et al. \shortcite{xiong2019modeling} have proposed automatically learning neural rewards that can encourage translation coherence at document level. However, it is not clear whether the learned rewards would be in good correspondence with human judgment. For this reason, in our work we prefer to rely on established discourse metrics as rewards.

\subsection{Discourse evaluation metrics}
\label{ssec:discourse_metrics}

As a matter of fact, several metrics have been proposed in the literature to measure discourse properties. For LC, Wong and Kit \shortcite{wong2012extending} have proposed a metric that looks for repetitions of words and their related terms (e.g. hyponyms, hypernyms) by using WordNet \cite{miller1998wordnet}. Gong et al. \shortcite{gong2015document} have proposed a similar metric that uses lexical chains. For COH, mainly two types of metrics have been proposed: \textit{entity-based} and \textit{topic-based}. The former follow the Centering Theory \cite{grosz1995centering} which states that documents with a high frequency of the same salient entities are more coherent. An entity-based coherence metric was proposed by Barzilay and Lapata \shortcite{barzilay2008modeling}. At their turn, \textit{topic-based} metrics assume that a document is coherent when adjacent sentences are similar in topic and vocabulary. Accordingly, Hearst \shortcite{hearst1997texttiling} has proposed the Texttiling algorithm which computes the cosine distance between the bag-of-word (BoW) vectors of adjacent sentences. Foltz et al. \shortcite{foltz1998measurement} have proposed to replace the BoW vectors with topic vectors. Li et al. \shortcite{li2016neural} have learned topic embeddings with a self-supervised neural network. There is also a third group of COH metrics that are based solely in syntactic regularities \cite{smith2016trouble} that have also shown to be effective at modelling textual coherence. Other metrics have been proposed to measure different discourse properties such as grammatical cohesion \cite{hardmeier2010modelling,miculicich2017validation} and discourse connectives \cite{hajlaoui2013assessing}.

\subsection{Reinforcement learning in NMT}
\label{ssec:rl_nmt}

Researchers in NMT and other natural language generation tasks have used reinforcement learning \cite{sutton2018reinforcement} techniques to train the models to maximize discrete sentence-level and document-level metrics as an alternative or a complement to the NLL. For example, Ranzato et al. \shortcite{ranzato2015sequence} have proposed training NMT systems targeting the BLEU score, showing consistent improvements with respect to strong baselines. In addition to training the model directly with the evaluation function, they claim that this approach mollifies the exposure bias problem \cite{bengio2015scheduled}. Expected risk minimization has been proposed as an alternative reinforcement learning-style training to maximize the sentence-level \cite{edunov2017classical,shen2015minimum} and the document-level \cite{saunders2020using} BLEU scores. Paulus et al. \shortcite{paulus2017deep} have proposed a similar approach for summarization using ROUGE as the training loss \cite{lin2000automated}. Tebbifakhr et al. \shortcite{tebbifakhr2019machine} have used a similar objective function to improve the sentiment classification of translated sentences. Finally, Edunov et al. \shortcite{edunov2017classical} have presented a comprehensive comparison of reinforcement learning and structured prediction losses for NMT model training.

\section{Baseline Models}
\label{sec:baseline}

This section describes the baseline NMT models used in the experiments. In detail, subsection \ref{ssec:standard} recaps the standard sentence-level translation model while subsection \ref{ssec:hierarchical} describes the recent, strong hierarchical baseline that we have augmented with discourse rewards.

\subsection{Sentence-level NMT}
\label{ssec:standard}

Our first baseline is a standard sentence-level NMT model. Given the source document $D$ with $k$ sentences, the model translates each sentence $\textbf{x}_i=\{\textbf{x}^1_i,\dots,\textbf{x}^{n_i}_i\}, i = 1,\ldots,k$, in the document into a  sentence in the target language, $\textbf{y}_i^\star=\{\textbf{y}^1_i,\dots,\textbf{y}^{m_i}_i\}$:

\begin{equation}
\label{eq:senNMT}
\textbf{y}^\star_{i}=\argmax_{\textbf{y}_i} p(\textbf{y}_i|\textbf{x}_i,\theta) \quad i=1,\dots,k 
\end{equation}

Thus, the model translates every sentence in the document independently. Our sentence model uses a standard transformer-based encoder-decoder architecture \cite{vaswani2017attention} where the model is trained to maximize the probability of the words in the training reference sentences using an NLL objective. We train this model for $20$ epochs and select the best model over the validation set. For more details on training and the hyper-parameters please see Appendix A.

\subsection{Hierarchical Attention Network}
\label{ssec:hierarchical}

As a document-level translation baseline, we have used the Hierarchical Attention Network (HAN) of Miculicich et al. \shortcite{miculicich2018document}. A HAN network is added to the sentence-level NMT model both in the encoder and in the decoder (referred to as HAN$_{\text{join}}$ in the following), allowing the model to encode information from $t$ previous source and target sentences. The prediction can be expressed as:

\begin{equation}
\label{eq:HAN_join}
\textbf{y}^\star_{i}=\argmax_{\textbf{y}_i} p(\textbf{y}_i|\textbf{x}_i,\textbf{x}_{i-1},\dots,\textbf{x}_{i-t},\textbf{y}_{i-1},\dots,\textbf{y}_{i-t},\theta) \quad i=1,\dots,k 
\end{equation}

\noindent where $(\textbf{x}_{i-1},\dots,\textbf{x}_{i-t})$ are the previous source sentences and $(\textbf{y}_{i-1},\dots,\textbf{y}_{i-t})$ the previous target sentences that make up the context. At inference time, the target sentences are the model's own predictions. Following the indications given by the authors, we have set $t=3$. Additionally, we have used the weights of the sentence-level NMT baseline to initialize the common parameters of the HAN$_{\text{join}}$ model, and we have initialized the extra parameters introduced by the HAN networks randomly. The model has been fine-tuned for $10$ epochs and the best model over the validation set has been selected. For further information on the hyper-parameters see Appendix A.

\begin{figure*}[t!]
	\centering
	\includegraphics[width=0.7\linewidth]{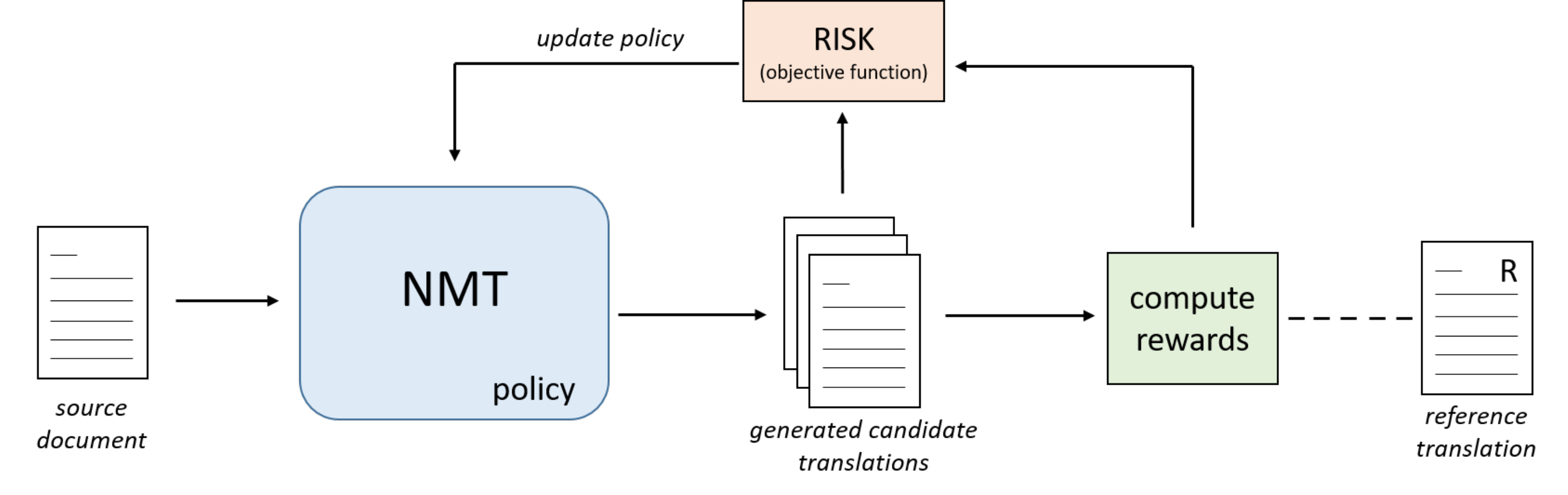}
	\caption{Risk training. Given the source document, the policy (NMT model) predicts $l$ candidate translations. Then, a reward function is computed for each such translation. For supervised rewards, (e.g., BLEU) the reference translation is required, but not for LC and COH. Finally, the Risk loss is computed using the rewards and the probabilities of the candidate translations, differentiated, and backpropagated for parameter update.}
	\label{fig:rl_diagram}
\end{figure*}

\section{Risk training with discourse rewards}
\label{sec:RISK}

In order to improve the baseline models, we propose to use the LC \cite{wong2012extending} and COH \cite{foltz1998measurement} evaluation metrics as rewards during training, so that the model is explicitly rewarded for generating more cohesive and coherent translation at document level. For that, we use a reinforcement learning approach, which allows using discrete, non-differentiable functions as rewards in the objective. Following Edunov et al. \shortcite{edunov2017classical}, we have used the structured loss that achieved the best results in their experiments, namely the \textit{expected risk minimization (Risk)} objective:

\begin{equation}
\label{eq:RISK}
\Lagr_{Risk} = \sum_{\textbf{u} \in \curlU(\textbf{x})} - r(\textbf{u},\textbf{y}) p(\textbf{u}|\textbf{x},\theta)
\end{equation}

\vspace{12pt}

\noindent where $\textbf{x}$ is the source sentence, $\textbf{y}$ is the reference translation, $p(\textbf{u}|\textbf{x})$ is the conditional probability of a translation in our `policy', or NMT model, $\curlU(x)$ is a set of candidate translations generated by the current policy, and $r(\cdot)$ is the reward function. In our work we have obtained the candidate translations using beam search, which achieved higher accuracy than sampling in Edunov et al. \shortcite{edunov2017classical}. The conditional probability of a translation has been defined as:

\begin{equation}
\begin{split}
\label{eq:cond_prob}
p(\textbf{u}|\textbf{x},\theta) &= \dfrac{f(\textbf{u},\textbf{x},\theta)}{\sum_{\textbf{u'} \in \curlU(\textbf{x})}f(\textbf{u'},\textbf{x},\theta)}\\
f(\textbf{u},\textbf{x},\theta) &= \exp[ \dfrac{1}{m}\sum_{j=1}^{m} \log p(u^j|u^1, \dots , u^{j-1},\textbf{x},\theta)]\\
\end{split}
\end{equation}

\vspace{12pt}

\noindent where $m$ is the number of words in the candidate translation. Note that in order to avoid underflow and put all the sentences on a similar scale, the (unnormalized) sentence score, $f(\textbf{u},\textbf{x},\theta)$, in Eq. \ref{eq:cond_prob} is computed as a sum of logarithms, divided by the number of tokens in the sequence and, finally, brought back to scale with the exponential function.

By minimizing this Risk objective, the NMT model is encouraged to give higher probability to candidate translations that obtain a higher reward. This function has been used at sentence level by Edunov et al. \shortcite{edunov2017classical}. However, the same metrics could also be computed at document level by simply concatenating all the sentences from the same document  together (both for the ground truth and the predictions). As a result, $m$ now would be the number of words in a document, $\curlU(\textbf{x})$ the candidate document translations, $\textbf{x}$ the source document and $\textbf{y}$ the reference document. Computing the Risk objective in this way permits having document-level reward functions as $r(\cdot)$. 

\subsection{Reward functions}
\label{ssec:reward}

We have explored the use and combination of different reward functions for training:

\vspace{8pt}

\textbf{$\textbf{LC}_{\textbf{doc}}$}: For LC, the metric proposed by Wong and Kit \shortcite{wong2012extending} has been adopted. This metrics counts the number of lexical cohesive devices in the document and then divides that number by the total number of words in the document (Eq. \ref{eq:lc}). Cohesive devices include associations such as repetitions of words, synonyms, near-synonyms, hypernyms, meronyms, troponyms, antonyms, coordinating terms, and so on. WordNet \cite{fellbaum2012wordnet} has been used to classify the relationships between words. Note that this reward function is unsupervised  since it does not require a ground-truth reference translation.

\begin{equation}
\label{eq:lc}
LC = \dfrac{\# \hspace{5pt} of \hspace{5pt} cohesion \hspace{5pt} devices \hspace{5pt} in \hspace{5pt} document}{\# \hspace{5pt} of \hspace{5pt} words \hspace{5pt} in \hspace{5pt} document}
\end{equation}

\vspace{8pt}

\textbf{$\textbf{COH}_{\textbf{doc}}$}: To calculate COH, we have used the approach proposed by Foltz et al. \shortcite{foltz1998measurement}. This approach first uses a trained LSA model to infer topic vectors ($\textbf{t}_i$) for each sentence in the document, and then computes the average cosine distance between adjacent sentences (Eq. \ref{eq:coh}). For the topic vectors, we have used the pre-trained LSA model (Wiki-6) from Stefanescu et al. \shortcite{stefanescu2014latent}, which was trained over Wikipedia. Note that COH also does not require a ground-truth reference translation.

\begin{equation}
\label{eq:coh}
COH = \frac{1}{k-1}\sum_{i=2}^{k} \cos(\textbf{t}_i,\textbf{t}_{i-1})
\end{equation}

\vspace{8pt}

$\textbf{BLEU}_{\textbf{doc}}$: In addition to the LC and COH rewards, we have decided to use a reference-based metric such as BLEU \cite{papineni2002bleus}. Due to the unsupervised nature of LC and COH, the model could trivially boost them by only repeating words and creating very similar sentences. However, this will come at the expense of producing translations that are increasingly unrelated to the reference translation (low adequacy) and grammatically incorrect (low fluency). As such, we encourage the model to also target a high BLEU score in its predictions.

\vspace{8pt}

\textbf{$\textbf{BLUE}_{\textbf{sen}}$}: Finally, we have also used BLEU at sentence level as a reward. In this way, we can assess whether it is more beneficial to use this metric at document or sentence level.

\vspace{12pt}

These four rewards can be combined in several different ways. To limit the experiments, we have decided to use them in their natural range without reweighting. All the results with the different reward combinations are presented in Section \ref{ssec:results}.

\subsection{Mixed objective}
\label{ssec:mixed}

Similar to the MIXER training proposed by Ranzato et al. \shortcite{ranzato2015sequence}, we have also explored mixing the Risk objective with the NLL. The rationale is similar to that of using BLEU$_{\text{doc}}$ and BLEU$_{\text{sen}}$ as rewards: the NLL loss can help the model to not deviate too much from the reference translation while improving discourse properties.
To mix these losses, we have used an alternate batch approach: either loss is randomly selected in each training batch, with a certain probability (e.g. Risk(0.8) means that we have selected the Risk loss with 80\% probability and the NLL with 20\%).

\section{Experiments}
\label{sec:experiments}

\subsection{Datasets and experimental setup}
\label{ssec:datasets}

\begin{table}[t]
	
	\begin{center}
		\resizebox{0.55\textwidth}{!}{\begin{tabular}{|l||c|c|c|c|c|}
				\hline
				\textbf{Language pair}&\textbf{Domain}&\textbf{train}&\textbf{dev}&\textbf{test}&\textbf{Avg. \# sent/doc}\\
				\hline\hline
				Zh-En&TEDtalks&0.2M&0.9K&3.9K&122\\
				Cs-En&TEDtalks&0.1M&0.5K&5.2K&114\\
				Es-En&TEDtalks&0.2M&0.8K&4.7K&114\\ 
				\hline
				Eu-En&subtitles&0.8M&0.8K&1.5K&1018\\
				Es-En&subtitles&1.1M&1.9K&4.6K&774\\
				\hline
				Es-En&news&0.2M&2.1K&14K&37\\
				\hline
		\end{tabular}}
		\caption{The datasets used for the experiments.}
		\label{tab_en_fr}
	\end{center}
\end{table}

We have performed a broad range of experiments over four different language pairs and three different translation domains (TED talks, movie subtitles and news) which have been used by other popular document-level NMT research \cite{miculicich2018document,tu2018learning}. For translations of TED talks\footnote{\href{https://wit3.fbk.eu/}{https://wit3.fbk.eu/}}, we have used the datasets released in the IWSLT14 for Spanish-English (Es-En), in the IWSLT15 shared task for Chinese-English (Zh-En) and in IWSLT16 for Czech-English (Cs-En). For both language pairs, we have used their \textit{dev2010} set as the validation set, and sets \textit{tst2011-2013} (Zh-En), \textit{tst2010-2013} (Cs-En) and \textit{tst2010-2012} (Es-En), respectively, as test sets. For translations in the movie subtitles domain, we have used the OpenSubtitles-v2018 dataset \cite{lison2019open} from OPUS\footnote{\href{http://opus.nlpl.eu/}{http://opus.nlpl.eu/}}, and the language pairs tested have been Basque-English (Eu-En) and Spanish-English (Es-En). For Eu-En we have used all the available data, but for Es-En we have only used a subset of the corpus to limit time and memory requirements. In both cases, we have divided the data into a training, validation and test sets\footnote{All the datasets will be released publicly, and the reviewers can already see them as supplementary material.}. The last translation domain is news, for which we have used the Es-En News-Commentary11 dataset\footnote{\href{http://www.casmacat.eu/corpus/news-commentary.html}{http://www.casmacat.eu/corpus/news-commentary.html}}. As validation and test sets, we have used its \textit{newstest2008} and \textit{newstest2009-2013} sets, respectively, from WMT\footnote{\href{http://www.statmt.org/wmt13/translation-task.html}{http://www.statmt.org/wmt13/translation-task.html}}. The document boundaries are given by the individual talks for the TED talks dataset, by movie scripts for the subtitles datasets and by single-author news commentaries for the news dataset.
All the datasets have been tokenized using the \textit{Moses tokenizer}\footnote{\href{https://github.com/moses-smt/mosesdecoder}{https://github.com/moses-smt/mosesdecoder}}, with the exception of Chinese for which we have used \textit{Jieba}\footnote{\href{https://github.com/fxsjy/jieba}{https://github.com/fxsjy/jieba}}. A \textit{truecaser} model from Moses$^7$ has been learned over the training data of each dataset, and has been applied for consistent word casing as a final pre-processing step.

As models, we have compared multiple models trained with the Risk objective with different combinations of reward functions. This has allowed us to select the best reward functions for the translation quality at document level. Then, the model trained with the best reward combination has been compared against the sentence-level NMT and HAN baselines. In our experiments, the Risk training objective has been used as fine-tuning of a pre-trained HAN$_{\text{join}}$ baseline model, in order not to suffer from a ``cold start'' due to the large output label space. The main aim of our experiments is to show that the proposed training objectives can lead to performance improvements over HAN$_{\text{join}}$.  Candidate translations have been obtained using beam search with a beam size of only 2, due to  memory and computational time limitations. Furthermore, the training batch size has been set to 15 sentences. Since the objective is computed over the batch, this is equivalent to subdividing longer documents into sub-documents of 15 sentences each. Yet, our experimental results show that computing the rewards at such batch level is still effective for improving the translation quality.

Each model has been trained with three different seeds over its training set, and the validation set has been used at all times to select the best model. Then, the average results of the three runs over the test set have been reported. We have measured four different evaluation metrics: BLEU, LC, COH and $\mathrm{F}_{\mathrm{BERT}}$, an alternative metric to BLEU that compares the BERT sentence embeddings of the prediction and the reference and which has been shown to have better correlation with the human judgement than BLEU \cite{zhang2019bertscore}. To select the best model over the validation set for the sentence-level NMT baseline, we have used the lowest perplexity. Instead, for the HAN$_{\text{join}}$ baseline and our models, we have chosen the model with the best results in the majority of the four evaluation metrics (BLEU, LC, COH and $\mathrm{F}_{\mathrm{BERT}}$). This has not affected the relative ranking of the sentence-level NMT baseline since its performance has been generally lower than the other approaches. Complete details about the experimental set-up and other hyper-parameters are provided in Appendix A. The code is publicly available\footnote{\url{https://github.com/ijauregiCMCRC/DL_NMT_RL}}.

\subsection{Results}
\label{ssec:results}

\begin{table}[t]	
	\begin{center}
		\begin{subtable}{\textwidth}
		\centering
		\resizebox{\textwidth}{!}{\begin{tabular}{|l||c c c c|c c c c|c c c c|}
				\hline
				\multirow{2}{*}{Model}&\multicolumn{4}{|c|}{\textbf{Zh-En (TED talks)}}&\multicolumn{4}{|c|}{\textbf{Cs-En (TED talks)}}&\multicolumn{4}{|c|}{\textbf{Es-En (TED talks)}}\\
				\cline{2-13}
				&BLEU&LC&COH&$\mathrm{F}_{\mathrm{\text{BERT}}}$&BLEU&LC&COH&$\mathrm{F}_{\mathrm{\text{BERT}}}$&BLEU&LC&COH&$\mathrm{F}_{\mathrm{\text{BERT}}}$\\
				\hline
				\hline
				Sentence-level NMT&16.94&55.39&28.02&66.94&22.74&55.62&27.72&69.60&39.55&56.67&28.27&79.5\\
				HAN$_{\text{join}}$&17.52&55.02&28.15&67.21&23.44&55.63&27.62&69.87&39.89&56.25&28.56&\textbf{79.88}\\
				\hdashline
				Human reference&--&55.13&29.33&--&--&55.91&29.7&--&--&57.84&30.79&--\\
				\hline
				Risk($1.0$)-$\mathrm{BLEU}_{\mathrm{doc}}+\mathrm{LC}_{\mathrm{doc}}+\mathrm{COH}_{\mathrm{doc}}$&\textbf{18.15}&\textbf{57.48}$^\ast$&\textbf{29.32}$^\ast$&67.69&23.40&\textbf{58.31}$^\ast$&\textbf{28.17}&\textbf{70.09}&37.4&59.41$^\dag$&28.92&78.86\\
				Risk($0.8$)-$\mathrm{BLEU}_{\mathrm{doc}}+\mathrm{LC}_{\mathrm{doc}}+\mathrm{COH}_{\mathrm{doc}}$&17.82&55.18&28.68&67.60&23.43&56.03$^\ast$&27.62&70.01$^\ast$&39.52&\textbf{57.53}&\textbf{28.79}&79.11\\
				Risk($0.5$)-$\mathrm{BLEU}_{\mathrm{doc}}+\mathrm{LC}_{\mathrm{doc}}+\mathrm{COH}_{\mathrm{doc}}$&17.83&54.70&28.30&\textbf{67.73}&23.42&56.07&27.78&69.95$^\ast$&\textbf{40.1}&57.4&28.78&79.61\\
				Risk($0.2$)-$\mathrm{BLEU}_{\mathrm{doc}}+\mathrm{LC}_{\mathrm{doc}}+\mathrm{COH}_{\mathrm{doc}}$&17.80&55.10&28.35&67.62&\textbf{23.48}&55.85&27.62&69.95&40.07&56.83&28.61&79.62\\
				\hline
		\end{tabular}}
		\caption{Results over the TED talks datasets.}
		\label{subtab:TEDtalks}
		\vspace{8pt}
		\end{subtable}
		\vspace{8pt}
		\begin{subtable}{\textwidth}
			\centering
			\resizebox{0.8\textwidth}{!}{\begin{tabular}{|l||c c c c|c c c c|}
					\hline
					\multirow{2}{*}{Model}&\multicolumn{4}{|c|}{\textbf{Eu-En (movie subtitles)}}&\multicolumn{4}{|c|}{\textbf{Es-En (movie subtitles)}}\\
					\cline{2-9}
					{}&BLEU&LC&COH&$\mathrm{F}_{\mathrm{\text{BERT}}}$&BLEU&LC&COH&$\mathrm{F}_{\mathrm{\text{BERT}}}$\\
					\hline
					\hline
					Sentence-level  NMT&9.12&37.08&19.34&59.18&29.34&58.31&22.70&67.57\\
					HAN$_{\text{join}}$&9.74&37.19&19.63&59.72&\textbf{30.14}&58.11&22.58&\textbf{67.73}\\
					\hdashline
					Human reference&--&41.83&21.93&--&--&57.28&24&--\\
					\hline
					Risk($1.0$)-$\mathrm{BLEU}_{\mathrm{doc}}+\mathrm{LC}_{\mathrm{doc}}+\mathrm{COH}_{\mathrm{doc}}$&1.19&72.51$^\dag$&27.67$^\dag$&36.72&3.37&67.82$^\dag$&19.53&48.07\\
					Risk($0.8$)-$\mathrm{BLEU}_{\mathrm{doc}}+\mathrm{LC}_{\mathrm{doc}}+\mathrm{COH}_{\mathrm{doc}}$&9.67&\textbf{40.66}$^\ast$&19.60&\textbf{59.76}&29.51&58.34&22.82&67.51\\
					Risk($0.5$)-$\mathrm{BLEU}_{\mathrm{doc}}+\mathrm{LC}_{\mathrm{doc}}+\mathrm{COH}_{\mathrm{doc}}$&9.77&38.85$^\ast$&\textbf{19.80}&59.62&29.79&\textbf{58.44}&22.76&67.53\\
					Risk($0.2$)-$\mathrm{BLEU}_{\mathrm{doc}}+\mathrm{LC}_{\mathrm{doc}}+\mathrm{COH}_{\mathrm{doc}}$&\textbf{9.99}&37.53&19.42&59.72&29.70&58.39&\textbf{22.96}&67.50\\
					\hline
			\end{tabular}}
			\caption{Results over the movie subtitles datasets.}
			\label{subtab:subtitles}
		\end{subtable}
		
		\begin{subtable}{\textwidth}
			\centering
			\resizebox{0.55\textwidth}{!}{\begin{tabular}{|l||c c c c|}
					\hline
					\multirow{2}{*}{Model}&\multicolumn{4}{|c|}{\textbf{Es-En (news)}}\\
					\cline{2-5}
					{}&BLEU&LC&COH&$\mathrm{F}_{\mathrm{\text{BERT}}}$\\
					\hline
					\hline
					Sentence-level NMT&21.79&32.97&28.1&67.88\\
					HAN$_{\text{join}}$&22.16&32.87&28.15&\textbf{68.28}\\
					\hdashline
					Human reference&--&38.66&30.97&--\\
					\hline
					Risk($1.0$)-$\mathrm{BLEU}_{\mathrm{doc}}+\mathrm{LC}_{\mathrm{doc}}+\mathrm{COH}_{\mathrm{doc}}$&20.67&32.81&28.14&67.84\\
					Risk($0.8$)-$\mathrm{BLEU}_{\mathrm{doc}}+\mathrm{LC}_{\mathrm{doc}}+\mathrm{COH}_{\mathrm{doc}}$&22.26&\textbf{33.70}$^\ast$&\textbf{28.45}$^\ast$&68.14\\
					Risk($0.5$)-$\mathrm{BLEU}_{\mathrm{doc}}+\mathrm{LC}_{\mathrm{doc}}+\mathrm{COH}_{\mathrm{doc}}$&22.34&33.51$^\ast$&28.39&68.02\\
					Risk($0.2$)-$\mathrm{BLEU}_{\mathrm{doc}}+\mathrm{LC}_{\mathrm{doc}}+\mathrm{COH}_{\mathrm{doc}}$&\textbf{22.45}$^\ast$&33.32$^\ast$&28.25&68.13\\
					\hline
			\end{tabular}}
			\caption{Results over the news datasets.}
			\label{subtab:news}
		\end{subtable}
		\caption{Main results. ($^\ast$) means that the differences are statistically significant with respect to the HAN$_{\text{join}}$ baseline with a p-value $<$ 0.05 over a one-tailed Welch's t-test. ($\dag$) indicates high LC and COH values that come at the expense of a considerable drop in translation accuracy (e.g. BLEU, $\mathrm{F}_{\mathrm{BERT}}$), and thus, likely undesirable.}
		\label{tab:results}
	\end{center}
\end{table}

Table \ref{tab:results} shows the main results from our experiments. Over all datasets, the HAN$_{\text{join}}$ baseline has consistently outperformed the sentence-level NMT in terms of BLEU score and $\mathrm{F}_{\mathrm{BERT}}$ which shows that including surrounding sentences can help to obtain better translation accuracy. However, HAN$_{\text{join}}$ has not performed significantly better than the sentence-level model in terms of LC and COH (even worse in a few cases), showing that it has not been able to specifically learn these discourse properties in the document. The COH and LC values of both  baselines have also been generally lower than those of the human reference translations for all datasets (with the exception of the LC in Zh-En (TED talks) and Es-En (movie subtitles)). 

Table \ref{tab:results} also shows the results from our best models in comparison to these baselines. From preliminary experiments, we have seen that the Risk model that achieved the best results is the one that combines BLEU${_\text{doc}}$, LC$_{\text{doc}}$ and COH$_{\text{doc}}$ as rewards. Yet, choosing the right proportion of Risk and NLL training has proven very important and dataset-dependent. In the TED talks domain (Table \ref{subtab:TEDtalks}), the Risk(1.0) model has outperformed the HAN$_{\text{join}}$ baseline in all evaluated metrics over the Zh-En dataset, improving $+0.63$ percentage points (pp) in BLEU, $2.46$ pp in LC, $1.17$ pp in COH and $0.48$ pp in $\mathrm{F}_{\mathrm{BERT}}$,  while in the Cs-En dataset the same model has got an improvement of $2.68$ pp in LC, $0.55$ pp in COH and $0.22$ pp in $\mathrm{F}_{\mathrm{BERT}}$, on a parity of BLEU score. Instead, over the Es-En dataset, even though the Risk(1.0) has achieved the highest LC and COH scores, this has come at a higher drop in translation accuracy (i.e. BLEU and $\mathrm{F}_{\mathrm{BERT}}$). Thus, we consider Risk(0.5) to be the best performing model over this dataset, as it still considerably improves LC and COH scores ($1.28$ pp and $0.23$ pp respectively), while keeping similar translation accuracy in terms of BLEU ($+0.22$ pp) and $\mathrm{F}_{\mathrm{BERT}}$ ($-0.27$ pp). In general, we had not anticipated the improvements in BLEU score and $\mathrm{F}_{\mathrm{BERT}}$ since our main aim had only been to improve the translations in terms of discourse metrics. However, in some cases the improvements in discourse metrics have also translated into higher translation accuracy.

In turn, Table \ref{subtab:subtitles} shows the main results over the movie subtitles datasets which are characterized by documents with, on average, more, yet much shorter, sentences than the TED talks. On these datasets, the Risk(1.0) model has been able to improve the LC and COH metrics to a large extent, but at a marked cost in BLEU score and $\mathrm{F}_{\mathrm{BERT}}$. Qualitatively, the translations generated by this model have often displayed many word and phrase repetitions that had little correspondence with the reference translation, showing that COH and LC can reach values that are undesirable. Conversely, training the model with the mixed objective has forced it to stay closer to the reference translations and helped it achieve higher BLEU and $\mathrm{F}_{\mathrm{BERT}}$ scores. On Eu-En, the Risk(0.8) model has improved the LC by $3.47$ pp at a substantial parity of all the other metrics. On Es-En, none of the proposed models has clearly outperformed the HAN$_{\text{join}}$ baseline. For instance, the Risk(0.5) model has improved LC and COH by $0.33$ pp and $0.18$ pp, respectively, but at the cost of $0.35$ pp in BLEU score and $0.20$ pp in $\mathrm{F}_{\mathrm{BERT}}$.

Finally, Table \ref{subtab:news} shows that the proposed models have delivered better results on the news domain dataset, where they have been able to simultaneously improve the BLEU score, LC and COH at a mild cost in $\mathrm{F}_{\mathrm{BERT}}$. In general, we can argue that the discourse rewards have proved more effective on documents such as talks and news commentaries -- which come from single authors and are generally controlled in style -- than on documents such as subtitles are more fragmented in nature.

\begin{table}[t]	
	\begin{center}
			\centering
			\resizebox{0.5\textwidth}{!}{\begin{tabular}{|l||c c c c|}
				\hline
				Model&BLEU&LC&COH&$\mathrm{F}_{\mathrm{\text{BERT}}}$\\
				\hline
				\hline
				$\mathrm{BLEU}_{\mathrm{doc}}+\mathrm{LC}_{\mathrm{doc}}+\mathrm{COH}_{\mathrm{doc}}$&\textbf{18.15}&57.48&29.32&67.69\\
				
				$\mathrm{BLEU}_{\mathrm{sen}}+\mathrm{LC}_{\mathrm{doc}}+\mathrm{COH}_{\mathrm{doc}}$&17.53&56.32&28.79&\textbf{67.96}\\
				\hline
				$\mathrm{BLEU}_{\mathrm{doc}}+\mathrm{LC}_{\mathrm{doc}}$&17.44&\textbf{59.21}&29.87&67.27\\
				
				$\mathrm{BLEU}_{\mathrm{doc}}+\mathrm{COH}_{\mathrm{doc}}$&17.57&55.74&28.82&67.41\\
				
				$\mathrm{BLEU}_{\mathrm{sen}}+\mathrm{LC}_{\mathrm{doc}}$&17.60&56.32&28.76&67.87\\
				
				$\mathrm{BLEU}_{\mathrm{sen}}+\mathrm{COH}_{\mathrm{doc}}$&17.46&56.31&28.82&67.93\\
				$\mathrm{LC}_{\mathrm{doc}}+\mathrm{COH}_{\mathrm{doc}}$&10.56&71.28$^\dag$&31.25$^\dag$&62.27\\
				\hline
				$\mathrm{BLEU}_{\mathrm{sen}}$&17.42&55.93&28.76&67.83\\
				$\mathrm{BLEU}_{\mathrm{doc}}$&17.20&54.59&28.15&67.18\\
				$\mathrm{LC}_{\mathrm{doc}}$&10.42&71.70$^\dag$&31.61$^\dag$&62.09\\
				$\mathrm{COH}_{\mathrm{doc}}$&17.26&58.66&\textbf{29.98}&66.92\\
				\hline
			\end{tabular}}
			\caption{Ablation study of the various reward functions over the Zh-En TED talks dataset with Risk(1.0). ($\dag$) indicates high LC and COH values that come at the expense of a considerable drop in translation accuracy (e.g. BLEU, $\mathrm{F}_{\mathrm{BERT}}$), and thus, likely undesirable.}
			\label{tab:ablation}
	\end{center}
\end{table}
\begin{table}[t]	
	\begin{center}
			\centering
			\resizebox{0.90\textwidth}{!}{\begin{tabular}{|l l|}
				\hline
				\textbf{Src}: &$\dots$ \begin{CJK*}{UTF8}{gbsn}女士们 ， 先生 们 ， 见见 你 的 近亲 。\end{CJK*}\\
				&\begin{CJK*}{UTF8}{gbsn}这 就是 \textcolor{green}{野生 倭 黑猩猩 的 世界   座落 于 刚果 的 丛林中} 。\end{CJK*}\\
				&\begin{CJK*}{UTF8}{gbsn}\textcolor{green}{倭 黑猩猩} 和 黑猩猩 是   我们 大家 生活 里 最 密切相关 的 近亲 。\end{CJK*}\\
				&\begin{CJK*}{UTF8}{gbsn}这 意味 我们 都 享有 一个 共同 的 祖先 ，\textcolor{green}{一个 进化 了 的 祖母 ，   她 生活 在 大约 6 百万年 前} 。\end{CJK*} $\dots$\\
				\hline
				\textbf{Ref}: &$\dots$ ladies and gentlemen , meet your cousins . \\
				&this is the world \textcolor{green}{of wild bonobos in the jungles of Congo} . \\
				&\textcolor{green}{bonobos} are , together with chimpanzees , your living closest relative . \\
				&that means we all share a common ancestor , \textcolor{green}{an evolutionary grandmother , who lived around six million years ago} $\dots$\\
				\hline
				\textbf{HAN$_{\text{join}}$}: &$\dots$ ladies and gentlemen , meet your relatives .\\
				&this is the world \textcolor{red}{of the wildlife that is in the Congo} .\\
				&\textcolor{red}{the chimps} and chimpanzees are the most closely related to us .\\
				&it means we all have a common ancestor , \textcolor{red}{a grandmother who has evolved about six million years ago} $\dots$\\
				\hdashline
				&\textbf{BLEU}: 32.21 \quad \textbf{LC}: 9.52 \quad \textbf{COH}:  19.36 \quad $\mathrm{\textbf{F}}_{\mathrm{\textbf{BERT}}}$:  80.78\\
				\hline
				\textbf{Risk(}$\textbf{1.0}$\textbf{)}:
				&$\dots$ ladies and gentlemen , meet your close relatives .\\
				&and that 's the world \textcolor{green}{of the wild bonobos that are in the jungle in the Congo} .\\
				&\textcolor{green}{the bonobos} are the most closely related to the chimpanzees that we live in .\\
				&it means that we all have a common ancestor , \textcolor{green}{a grandmother who lived about six million years ago} $\dots$\\
				\hdashline
				&\textbf{BLEU}: 23.99 \quad \textbf{LC}: 20.00 \quad \textbf{COH}:  20.77 \quad $\mathrm{\textbf{F}}_{\mathrm{\textbf{BERT}}}$:  82.83\\
				\hline
			\end{tabular}}
	\caption{Translation example. Excerpt of a document from the Zh-En TED talks test set.}
	\label{tab:example}
    \end{center}
\end{table}

\subsubsection{Ablation study and translation example}
\label{sssection:ablation}

To expand the analysis, Table \ref{tab:ablation} shows the results from an ablation study that explores the impact of the various reward functions over the Zh-En dataset. The best trade-off over the four evaluation metrics seems that returned by BLEU$_{\text{doc}}$ + LC$_{\text{doc}}$ + COH$_{\text{doc}}$ which has achieved the highest BLEU score, a high $\mathrm{F}_{\mathrm{BERT}}$, and high LC and COH. The results also show that using BLEU$_{\text{sen}}$ as a reward has contributed to improve the $\mathrm{F}_{\mathrm{BERT}}$ score in all cases, but at the significant expense of the other evaluation metrics. However, when BLEU$_{\text{doc}}$ and BLEU$_{\text{sen}}$ have been compared head-to-head as the sole rewards, the sentence-level BLEU has been able to achieve higher scores in all metrics. In contrast,  the BLEU$_{\text{doc}}$ reward has been most effective when used jointly with the cohesion and coherence rewards. At its turn, the LC$_{\text{doc}}$ reward without a balance from a BLEU reward has led to LC and COH scores that are likely excessive and undesirable, with a corresponding drop in BLEU score and $\mathrm{F}_{\mathrm{BERT}}$. Conversely, the COH$_{\text{doc}}$ reward has not displayed a comparable degradation. The main overall result from this ablation analysis is that the rewards need to be used in a calibrated combination to deliver the best trade-off across all the evaluation metrics, and that the selection of the best combination can be effectively carried out by validation.


Finally, Table \ref{tab:example}  shows an example of the translation of a document excerpt from the Zh-En TED talks dataset made by our best model (Risk(1.0)-$BLEU_{doc}+LC_{doc}+COH_{doc}$), in comparison to that made by the HAN$_{\text{join}}$ baseline, the reference translation (Ref) and the text in the source language (Src). In this example, we can clearly see the positive influence of the LC and COH rewards, as the model has been able to provide better lexical cohesion and coherence in the translation. The model has also been able to correctly translate words such as \textit{bonobos} and \textit{jungle} while the HAN$_{\text{join}}$ model has uttered a more generic \textit{chimps}. In addition, the translation generated by our model seems more faithful to the reference overall. Note also that these improvements have come at a significant drop in BLEU score. This may suggest that LC and COH can influence improvements that the BLEU score is not able to capture. Examples for the other language pairs are provided in Appendix B.


\section{Conclusion}
\label{sec:conclusion}

In this paper, we have presented a novel training method for document-level NMT models that uses discourse rewards to encourage the models to generate more lexically cohesive and coherent translations at document level. As training objective we have used a reinforcement learning-style function, named Risk, that permits using discrete, non-differentiable terms in the objective. Our results on four different language pairs and three translation domains have shown that our models have achieved a consistent improvement in discourse metrics such as LC and COH, while retaining comparable values of accuracy metrics such as BLEU and $\mathrm{F}_{\mathrm{BERT}}$. In fact, on certain datasets, the models have even improved on those metrics. While the approach has proved effective in most cases, the best combination of discourse rewards, accuracy rewards and NLL has had to be selected by validation for each dataset. In the near future we plan to investigate how to automate this selection, and also explore the applicability of the proposed approach to other natural language generation tasks.

\section*{Acknowledgment}
\label{acknowledgment}

The   authors   would   like   to   thank  the RoZetta Institute (formerly CMCRC) for providing  financial  support to this research. Warmest thanks also go to Dr. Sameen Maruf for her feedback on an early version of this paper.

\bibliographystyle{acl}
\bibliography{bibliography}

\newpage

\appendix

\section*{Appendix A: Training and hyper-parameters}
\label{appendix_a}

For our experiments, we have used, and developed on top of, the code provided by Miculicich et al. \shortcite{miculicich2018document} based on the OpenNMT framework \cite{klein2017opennmt}. Our code is publicly available\footnote{\url{https://github.com/ijauregiCMCRC/DL_NMT_RL}}.

\textbf{Sentence-level NMT}: For the sentence-level model, we have used the hyper-parameters proposed by Miculicich et al. \shortcite{miculicich2018document} in their code repository\footnote{\href{https://github.com/idiap/HAN\_NMT}{https://github.com/idiap/HAN\_NMT}}. The model uses a $6$-layer transformer network \cite{vaswani2017attention} as the encoder and decoder. The dimensions of the source word embeddings, the target word embeddings and the transformers' hidden vectors have all been set to $512$. The default position encoding has been added to the input vectors, and a dropout of $0.1$ to the hidden vectors. Additionally, a label smoothing of $0.1$ has been applied to the output probabilities. During training, the batch size has been set to $4096$ tokens with a gradient accumulation of $4$. We have used the Adam optimizer \cite{kingma2014adam} with a learning rate of $2$, and $\beta_2=0.998$. The parameters of the network have been initialized with the \textit{glorot} method \cite{glorot2010understanding}, and the model has been warmed up for $8,000$ steps. Training has been performed for $20$ epochs and the model with the best perplexity over the validation set has been selected.

\textbf{HAN$_{\text{join}}$}: The document-level baseline follows almost exactly the settings of the sentence-level one. The main difference is the added HAN networks in the encoder and decoder. During training, for memory reasons the batch size has been reduced to $1,024$ tokens, and the learning rate to $0.2$. The parameters in common have been initialized with the pre-trained sentence-level baseline, while the extra HAN networks have been initialized with \textit{glorot}. For computational reasons, this model has been trained for only $10$ epochs. In the validation, the best model was selected as that with the highest values in the majority of the evaluation metrics (BLEU, LC, COH and $\mathrm{F}_{\mathrm{BERT}}$).

\textbf{Risk models}: All our proposed models use the HAN$_{\text{join}}$ architecture. As such, they all have been initialized with the pre-trained weights of the HAN$_{\text{join}}$ baseline, and then fine tuned with the Risk objective. The candidate documents for the Risk objective have been obtained with beam search and limited to just $2$, due to limitations in computational resources. For the same reason, the batch size has had to be limited to 15. This has led to splitting most of the documents into multiple batches, and the reward metrics have been computed at batch level. For the batches that contained a document boundary, we have computed the rewards separately for each document. The model has been fine-tuned until convergence of the perplexity on the validation set, and using simulated annealing \cite{denkowski2017stronger}, which repeatedly halves the learning rate when perplexity convergence is reached. The number of annealing steps has been set to $5$. After training, the model with the highest values in the majority of the evaluation metrics (BLEU, LC, COH and $\mathrm{F}_{\mathrm{BERT}}$) has been selected.

\section*{Appendix B: Translation examples}
\label{appendix_b}

In this appendix we show other translation examples that give evidence to the translation improvement achieved by our models. Table \ref{tab:app_example_1} shows a Cs-En translation example. The example shows that our model has successfully translated word \textit{renewables} while the baseline predicted \textit{electricity}. Additionally, it has properly constructed the phrase \textit{to build a completely new system}. Table \ref{tab:app_example_2} shows an Eu-En example, where our model has properly predicted word \textit{card} instead of the baseline's \textit{ticket}. Our model has also predicted sentence \textit{they 'll lock me in a mental hospital} which, by looking at the source excerpt, seems more adequate than the translation provided by the baseline, and even possibly the reference sentence itself. Table \ref{tab:app_example_3} shows an Es-En example in the news domain. Our model has predicted phrase \textit{any late consequence can be avoided}, which, again, seems more appropriate than the baseline's prediction \textit{any belated consequence is possible}. Finally, Table \ref{tab:app_example_4} shows how our model seems to have better captured the context of the excerpt, which revolves around money and payments, and has correctly translated the Spanish word \textit{adelanto} for \textit{advancement}. Conversely, the translation from the HAN$_{\text{join}}$ baseline has been \textit{earlier}, which could be correct in a different context, but not in this one.

\begin{table}[h]	
	\begin{center}
		\centering
		\resizebox{0.95\textwidth}{!}{\begin{tabular}{|l l|}
				\hline
				\textbf{Src}: & $\dots$ ot\'{a}zka zn\'{i} :  `` m\r{u}\u{z}eme ho sn\'{i}\u{z}it na nulu ? '' \\
				&pokud budeme spalovat uhl\'{i} , tak ne . \\
				&ani p\u{r}i spalov\'{a}n\'{i} zemn\'{i}ho plynu ne . \\
				&t\'{e}m\u{e}\u{r} ka\u{z}d\'{y} sou\u{c}asn\'{y} zp\r{u}sob v\'{y}roby elekt\u{r}iny , s vyj\'{i}mkou roz\u{s}i\u{r}uj\'{i}c\'{i}ch \textcolor{green}{se obnoviteln\'{y}ch a jadern\'{y}ch zdroj\r{u}} , produkuje CO2 .\\
				&budeme muset v glob\'{a}ln\'{i}m m\u{e}\u{r}\'{i}tku \textcolor{green}{vytvo\u{r}it \'{u}pln\u{e} nov\'{y} syst\'{e}m} .\\
				&a pot\u{r}ebujeme energetick\'{e} z\'{a}zraky $\dots$\\
				\hline
				\textbf{Ref}: &$\dots$ and so the question is : can you actually get that to zero ?\\
				&if you burn coal , no .\\
				&if you burn natural gas , no .\\
				&almost every way we make electricity today , except for the emerging \textcolor{green}{renewables and nuclear} , puts out CO2 .\\
				&and so , what we ’re going to have to do at a global scale , \textcolor{green}{is create a new system} .\\
				&and so , we need energy miracles $\dots$\\
				\hline
				\textbf{HAN$_{\text{join}}$}: &$\dots$ the question is , can we reduce it to zero ?\\
				&if we keep burning coal , we don 't .\\
				&even burning , natural gas don ’t .\\
				&almost every single way of production of electricity , except for the exception of \textcolor{red}{electricity} and nuclear resources , produces CO2 . \\
				&we ’re going \textcolor{red}{to have to have a completely new system} on a global scale .\\
				&and we need energy miracles $\dots$\\
				\hline
				\textbf{Risk(}$\textbf{1.0}$\textbf{)-}&$\dots$ the question is , can we reduce it to zero ?\\
				&if we keep burning coal , we don ’t .\\
				&even burning , natural gas don ’t .\\
				&almost every single way of producing electricity , except for example , with the exception \textcolor{green}{of renewables and nuclear resources} , produces CO2 .\\
				& we ’re going to have \textcolor{green}{to build a completely new system} on a global scale .\\
				&and we need energy miracles $\dots$\\
				\hline
		\end{tabular}}
		\caption{Translation example. Excerpt of a document from the Cs-En TED talks test set.}
		\label{tab:app_example_1}
	\end{center}
\end{table}

\begin{table}[h]	
	\begin{center}
		\centering
		\resizebox{0.5\textwidth}{!}{\begin{tabular}{|l l|}
				\hline
				\textbf{Src}: & $\dots$ ` zure izena ? \\
				&\textcolor{green}{baduzu txartelik} ? '\\
				&` zure helbidea ?\\
				&baduzu telefonorik etxean ? '\\
				&`  eskerrik asko . joan zaitezke . \\
				&harremanetan egongo gara ' .\\
				&` nire familiak hau jakiten badu , \textcolor{green}{eroetxe batean giltzapetuko naute} '$\dots$\\
				\hline
				\textbf{Ref}: &$\dots$ ` your name ?\\
				&\textcolor{green}{do you have a card} ? '\\
				&` your address ?\\
				&do you have a telephone in your home ? '\\
				&` thank you , that ’s fine .\\
				&you can leave . I will contact you later . '\\
				&` if my family learns about this , \textcolor{red}{I will be forcefully detained} . ' $\dots$\\
				\hline
				\textbf{HAN$_{\text{join}}$}: &$\dots$ your name ?\\
				&\textcolor{red}{do you have a ticket} ? '\\
				&your address ?\\
				&do you have a phone at home ? '\\
				&` thank you .\\
				&we 'll be in touch . '\\
				&` if my family knows this , \textcolor{red}{I 'll be locked up} \textcolor{green}{in a mental institution} . '$\dots$\\
				\hline
				\textbf{Risk(}$\textbf{1.0}$\textbf{)-}&$\dots$ ' your name ?\\
				&\textcolor{green}{do you have a card} ? '\\
				&your address ?\\
				&do you have a phone at home ? '\\
				&thank you .\\
				&we 'll be in touch . '\\
				&` if my family knows this , \textcolor{green}{they 'll lock me in a mental hospital} . ' $\dots$\\
				\hline
		\end{tabular}}
		\caption{Translation example. Excerpt of a document from the Eu-En subtitles test set.}
		\label{tab:app_example_2}
	\end{center}
\end{table}

\newpage

\begin{table}[h]	
	\begin{center}
		\centering
		\resizebox{\textwidth}{!}{\begin{tabular}{|l l|}
				\hline
				\textbf{Src}: & $\dots$ los preservativos pueden reducir el riesgo de contagio , pero no ofrecen protecci\'{o}n al cien por cien . \\
				& algunos agentes patógenos de enfermedades de transmisi\'{o}n sexual tambi\'{e}n pueden transmitirse a trav\'{e}s de infecciones por suciedad y por contacto f\'{i}sico .\\
				&por este motivo , los expertos recomiendan someterse regularmente a ex\'{a}menes m\'{e}dicos , sobre todo si se cambia con frecuencia de pareja sexual . \\
				&si se diagnostican de forma temprana , la mayor\'{i}a de las ETS pueden curarse y \textcolor{green}{es posible evitar cualquier consecuencia tard\'{i}a} $\dots$\\
				\hline
				\textbf{Ref}: &$\dots$ condoms can reduce the risk of contraction , however , they do not offer 100 \% protection .\\
				&this is because occasionally , the pathogens of sexually transmitted diseases can also be passed on via smear infections and close bodily contact .\\
				&therefore , first and foremost experts recommend that people with frequently changing sexual partners undergo regular examinations .\\
				&if diagnosed early , the majority of STIs can be cured and \textcolor{green}{long \@-\@ term consequences avoided} $\dots$\\
				\hline
				\textbf{HAN$_{\text{join}}$}: &$\dots$ condoms can reduce the risk of contagion , but they do not provide protection to 100 per hundred .\\
				&some $<$unk$>$ agents of sexual transmission can also be cured by $<$unk$>$ infections and physical contact .\\
				&for this reason , experts recommend submitting regularly to medical tests , especially if sexual couple are often changed .\\
				&if taken early , most of the ETS can collapse and \textcolor{red}{any belated consequence is possible} $\dots$\\
				\hline
				\textbf{Risk(}$\textbf{1.0}$\textbf{)-}&$\dots$ condoms can reduce the risk of contagion , but they do not provide protection to a hundred per hundred .\\
				&some immune agents of sexual transmission can also be channeled through infection by $<$unk$>$ and physical contact .\\
				&for this reason , experts report regularly to medical tests , especially if sexual couple are often changed .\\
				&if they were early in , most of the ETS can collapse , and \textcolor{green}{any late consequence can be avoided} $\dots$\\
				\hline
		\end{tabular}}
		\caption{Translation example. Excerpt of a document from the Es-En news test set.}
		\label{tab:app_example_3}
	\end{center}
\end{table}

\begin{table}[h]	
	\begin{center}
		\centering
		\resizebox{0.90\textwidth}{!}{\begin{tabular}{|l l|}
				\hline
				\textbf{Src}: & $\dots$ no voy a perdonar a ese bastardo ! \\
				&Digaselo al Dr Chaddha , no me mienta . \\
				&le dí el \textcolor{green}{30 \% por adelantado} $\dots$ \\
				&incluso después de haberme prometido , que él nos daria una esperma de calidad . digale que se joda !$\dots$\\
				\hline
				\textbf{Ref}: &$\dots$ I wont spare that bas****\\
				&tell that Dr. Chaddha of yours , not to lie to me .\\
				&he has taken \textcolor{green}{30 \% advance from me} $\dots$\\
				&even after promising , he hasn ’t given us a quality sperm . you tell that f * * * er I ’ll hunt him down$\dots$\\
				\hline
				\textbf{HAN$_{\text{join}}$}: &$\dots$ I 'm not going to forgive that bastard !\\
				&don 't lie to me .\\
				&I gave him \textcolor{red}{30 \% earlier} .\\
				&even after I was promised , he 'd give us a quality sperm$\dots$\\
				\hline
				\textbf{Risk(}$\textbf{0.8}$\textbf{)}:
				&$\dots$ I 'm not going to forgive that bastard !\\
				&don 't lie to me .\\
				&I gave him \textcolor{green}{30 \% advance} .\\
				&even after I was promised , he 'd give us a quality sperm $\dots$\\
				\hline
		\end{tabular}}
		\caption{Translation example. Excerpt of a document from the Es-En subtitles test set.}
		\label{tab:app_example_4}
	\end{center}
\end{table}

\section*{Appendix C: Rewards during training}
\label{appendix_c}

\begin{figure*}[h]
	\centering
	\includegraphics[width=0.7\linewidth]{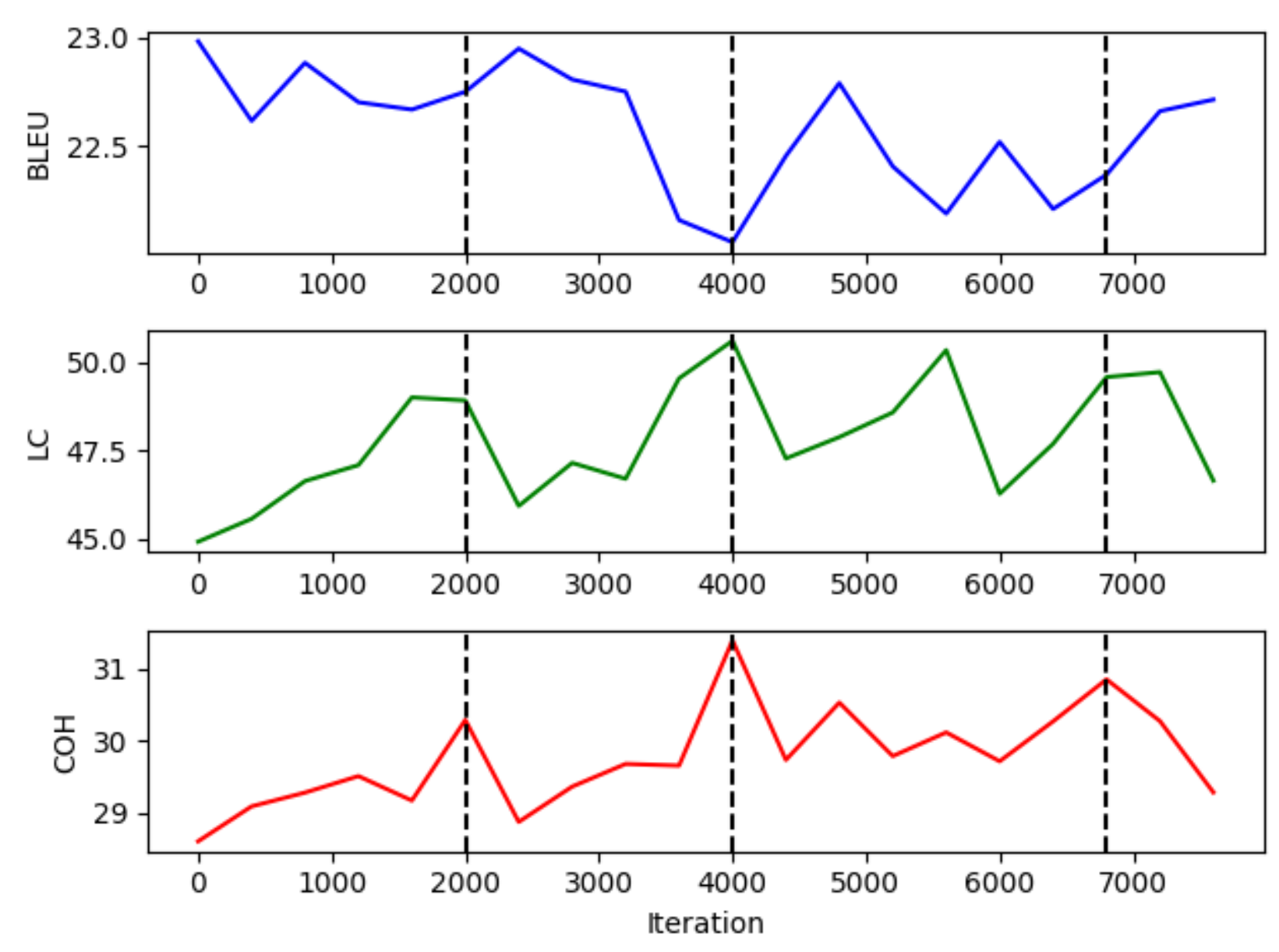}
	\caption{BLEU, LC and COH scores over the Cs-En validation set at different training iterations.}
	\label{fig:rewards}
\end{figure*}

To show the behavior of the different rewards during training, Figure \ref{fig:rewards} shows the BLEU, LC and COH scores over the Cs-En validation set at different training iterations. This plot confirms the intuition that improving LC and COH comes at a cost of BLEU score. In the first 2000 training iterations, LC has improved by more than 2 pp and COH by more than 1 pp, while the BLEU score has dropped by approximately 0.3 pp. Moreover, the highest scores for LC and COH coincide with the lowest score for BLEU (iteration 4000). Overall, validation is needed to achieve a model with the best trade-off between BLEU, LC and COH (in this case, for instance, iteration 2000 or 6800).

\end{document}